\documentclass[conference]{IEEEtran}
\IEEEoverridecommandlockouts
\usepackage{cite}
\usepackage{amsmath,amssymb,amsfonts}
\usepackage{algorithmic}
\usepackage{graphicx}
\usepackage{textcomp}
\usepackage{xcolor}
\usepackage{graphicx}
\usepackage{balance}

\usepackage{amsmath}
\usepackage{amsthm}
\usepackage{booktabs}
\usepackage{algorithm}
\usepackage{algorithmic}
\usepackage{multirow}
\usepackage{subfigure}
\def\BibTeX{{\rm B\kern-.05em{\sc i\kern-.025em b}\kern-.08em
    T\kern-.1667em\lower.7ex\hbox{E}\kern-.125emX}}
\begin{document}

\title{Trusted Mamba Contrastive Network for Multi-View Clustering}

\author{
Jian Zhu$^{1,3,\dagger}$,
Xin Zou$^{2,\dagger}$,
Lei Liu$^{1,\ast}$,
Zhangmin Huang$^{3}$,
Ying Zhang$^{3}$,
Chang Tang$^{2}$,
Li-Rong Dai$^{1}$
\thanks{$\dagger$ Contributed equally to this work.}
\thanks{$\ast$ Corresponding author.}\\
\IEEEauthorblockA{
$^{1}$University of Science and Technology of China, 
\{liulei13, lrdai\}@ustc.edu.cn\\
$^{2}$China University of Geosciences, 
\{zouxin, tangchang\}@cug.edu.cn\\
$^{3}$Zhejiang Lab, \{qijian.zhu, zmhuang, yingzhang\}@zhejianglab.com
}
}

\maketitle

\begin{abstract}
Multi-view clustering can partition data samples into their categories by learning a consensus representation in an unsupervised way and has received more and more attention in recent years. However, there is an untrusted fusion problem. The reasons for this problem are as follows: 1) The current methods ignore the presence of noise or redundant information in the view; 2) The similarity of contrastive learning comes from the same sample rather than the same cluster in deep multi-view clustering. It causes multi-view fusion in the wrong direction. This paper proposes a novel multi-view clustering network to address this problem, termed as Trusted Mamba Contrastive Network (TMCN). Specifically, we present a new Trusted Mamba Fusion Network (TMFN), which achieves a trusted fusion of multi-view data through a selective mechanism. Moreover, we align the fused representation and the view-specific representation using the Average-similarity Contrastive Learning (AsCL) module. AsCL increases the similarity of view presentation from the same cluster, not merely from the same sample. Extensive experiments show that the proposed method achieves state-of-the-art results in deep multi-view clustering tasks. The source code is available at https://github.com/HackerHyper/TMCN.
\end{abstract}


\section{Introduction}
With the rapid growth of digitization, data is collected from various views. For instance, autonomous driving systems integrate data from multiple cameras to make decisions \cite{chen2024end}. 
The term ``multi-view data" refers to an object that is represented from multiple perspectives~\cite{zou2023dpnet}. Multi-view clustering (MVC)~\cite{chao2021survey,zou2023inclusivity,xiao2024dual,yang2024trustworthy} seeks to fuse these diverse views to identify meaningful groupings unsupervised, making it crucial to data mining \cite{dang2024exploring,he2023multispectral,xu2025hstrans,zou2024dai}. However, this remains a challenging problem.

\begin{figure*}
  \centering
  \includegraphics[width=18cm]{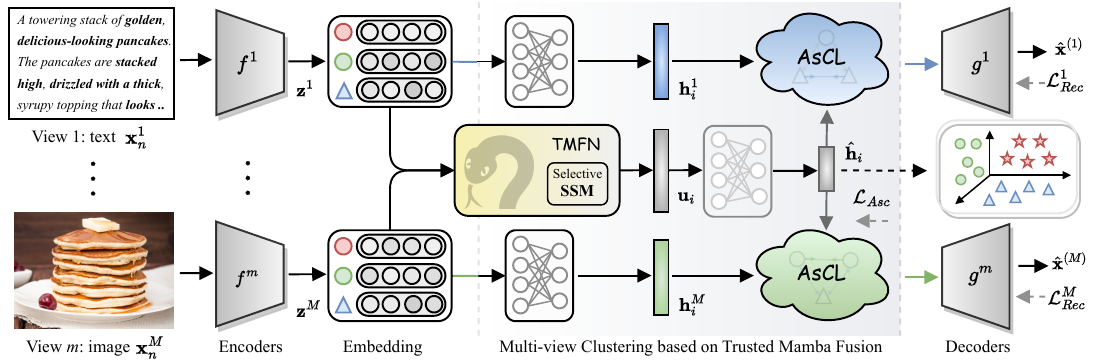}
  \caption{Overall Framework of TMCN. The framework consists of TMFN and AsCL. TMFN segments the one-dimensional feature vector of each view into sequence vectors, followed by a trusted fusion of multi-view features via a selective mechanism of the Mamba network. In contrast, AsCL is introduced to enhance the similarity of view representations within the same cluster, rather than merely focusing on the similarity at the individual sample level. It further improves the trusted fusion of multi-view data.}
  \label{fig:02}
\end{figure*}

In natural language processing tasks, deep learning has demonstrated outstanding effectiveness in data representation~\cite{zou2023hierarchical, zhu:23, zhu:24, zhu:25, zhu:26, zou2024look, zhu:27}. Similarly, deep clustering has also seen significant advancements~\cite{du2021deep, abavisani2018deep, zhou2020end, xu2021deep, trosten2021reconsidering, xu2022multi}. These methods leverage a view-specific encoder network to generate effective embeddings, which are then combined from all views for deep clustering. To mitigate the impact of view-specific information on clustering, several alignment networks have been proposed. For instance, some approaches utilize KL divergence to align the label or representation distributions across multiple views~\cite{hershey2007approximating}. However, the fact that different views of a sample may belong to different categories presents a significant challenge to deep clustering. To address this, certain methods employ contrastive learning to align representations from various views.

Even though these methods have made substantial progress in addressing the MVC challenge, the issue of untrusted fusion persists. This problem arises for several reasons: 1) A view or multiple views of a sample may contain excessive noise or redundant data. Generating a reliable representation from multiple views is challenging because nearly all deep MVC methods (e.g., CoMVC \cite{trosten2021reconsidering}, DSIMVC \cite{tang2022deep}, DIMVC \cite{xu2022deep}) rely on simple fusion techniques, such as weighted-sum fusion or concatenation of all views. 2) At the sample level, alignment methods based on contrastive learning (e.g., MFLVC \cite{xu2022multi}, DSIMVC \cite{tang2022deep}) typically differentiate between positive and negative pairs. However, this approach may conflict with the clustering objective, which requires that representations within the same cluster be similar. The contrastive learning loss can cause the fused representation to drift in the wrong direction, leading to untrusted fusion. These factors ultimately reduce the performance of multi-view clustering.

We propose a Trusted Mamba Contrastive Network (TMCN) for clustering to address the aforementioned issues. Inspired by the outstanding features of the Mamba network \cite{Albert:28}, TMCN leverages a selection mechanism to learn trusted representations from multi-view data. Additionally, to overcome clustering challenges, we enhance the similarity of view representations from the same cluster in contrastive learning, rather than focusing solely on the same sample. To achieve trusted fusion, we first use an autoencoder model to obtain view-specific representations that effectively reconstruct the original data. We then introduce a Trusted Mamba Fusion Network (TMFN) to perform reliable fusion of the multi-view data. Finally, we propose Average-similarity Contrastive Learning (AsCL) to enhance the similarity of view representations within the same cluster, rather than limiting it to the same sample. Our main contributions are summarized as follows:
\begin{itemize}
\item We propose TMFN for deep multi-view clustering, which implements multi-view trusted fusion through the filtration capabilities of the Mamba selection mechanism. To the best of our knowledge, we are the first to utilize the selection mechanism for multi-view trusted fusion.
\item Moreover, we introduce AsCL scheme to enhance intra-cluster view-specific representation similarity, in contrast to previous approaches that treat different views of an instance as positive samples, to facilitate trusted fusion.
\item Experimental results demonstrate that our proposed TMCN achieves state-of-the-art performance in deep multi-view clustering tasks on various datasets.
\end{itemize}

\section{The Proposed Methodology}
\label{section:proposed_method}
We propose an innovative TMCN, which aims to solve untrusted fusion of multi-view data. The proposed TMCN is shown in Fig. \ref{fig:02}. It mainly consists of three components: 1) Multi-view Data Reconstruction, 2) Trusted Mamba Fusion Network, and 3) Average-similarity Contrastive Learning. The multi-view data, which includes $N$ samples with $M$ views, is denoted as $\{\mathbf{X}^{m}=\{{x}_{1}^{m};...;{x}_{N}^{m}\}\in \mathbb{R}^{N \times D_{m}}\}_{m=1}^M$, where $D_m$ is the feature dimension in the $m$-th view. 

\subsection{Multi-view Data Reconstruction}
We use Autoencoder\cite{song2018self,hinton2006reducing} to extract individual view features. It has two parts: an encoder and a decoder. The encoder function is denoted by $f^{m}$ for the $m$-th view. The encoder generates the low-dimensional embedding as follows:
\begin{equation}
\label{eq:encoder}
\begin{aligned}
{z}_{i}^{m}=f^{m}\left({x}_{i}^{m}\right), 
\end{aligned}
\end{equation} 
where ${z}_{i}^{m} \in \mathbb{R}^{d_m}$ is the embedding of the $i$-th sample from the $m$-th view ${x}_{i}^{m}$. $d_m$ is the dimension of the feature.

Using the data representation ${z}_{i}^{m}$, the decoder reconstructs the sample. Let $g^{m}$ denote the decoder function. In the decoder component, ${z}_{i}^{m}$ is decoded to provide the reconstructed sample ${\hat{x}}_{i}^{m}$:
\begin{equation}
\label{eq:decoder}
\begin{aligned}
\hat{{x}}_{i}^{m}=g^{m}\left({z}_{i}^{m}\right).
\end{aligned}
\end{equation}

Let $\mathcal{L}_{Rec}$ represent the reconstruction loss, $N$ denotes the number of
samples. The following formula is used to calculate the reconstruction loss
\begin{equation}
\label{eq:reconstruction loss}
\begin{aligned}
\mathcal{L}_{\mathrm{Rec}}=&\sum_{m=1}^{M}\left\|{X}^{m}-\hat{{X}}^{m}\right\|_{2}^{2}\\
=&\sum_{m=1}^{M}\sum_{i=1}^{N}\left\|{x}_i^{m}-g^{m}\left({z}_{i}^{m}\right)\right\|_{2}^{2}.
\end{aligned}
\end{equation}

\subsection{Trusted Mamba Fusion Network}\label{sec:ssm}
We propose TMFN to implement the trusted fusion of multi-view data. It consists of three modules: Fine-grained Network, Mamba Network, and Convert Network.

\noindent \textbf{Fine-grained Network.} We first transform the one-dimensional feature vector of each view-specific sample into a detailed sequence vector, as follows:
\begin{equation}
e^m_i=rea^{1}(z^m_i), e^m_i \in \mathbb{R}^{l \times d},
\end{equation}
where $l$ represents the length of the sequence vector, and $d$ represents the dimension of the sequence vector. $rea$ is the operation of fine-grained segmentation of sequences.
The sequence vectors of each view will be concatenated together to form a global sequence vector as follows:
\begin{equation}
e_i=cat(e^1_i, e^2_i \dots, e^M_i), e_i \in \mathbb{R}^{Ml \times d}.
\end{equation}

\noindent \textbf{Mamba Network.} We employ Mamba's selection mechanism for the trusted fusion of multi-view data. The network consists of two distinct branches as follows:
\begin{equation}
p_i=mlp^1(e_i), ~~q_i=mlp^2(e_i)~~(p_i, q_i\in \mathbb{R}^{Ml \times d^{'}}),
\end{equation}
where $mlp$ represents multi-layer perceptron networks. Here, we use MLP to upscale the original sequence vector $e_i$. $d^{'}$ represents the dimension of the sequence vector after expansion ($d^{'}=d*\alpha$). $\alpha$ represents the coefficient of expansion.
\begin{equation}
p{'}_i=rea^{3}(conv1d(rea^{2}(p_i))), p{'}_i \in \mathbb{R}^{d^{'} \times Ml},
\end{equation}
where $conv1d$ denotes a one-dimensional convolutional neural network. The selection mechanism is formulated as follows:
\begin{equation}
\begin{aligned}
\label{eq:discret-ssm}
    & h_k = \overline{\mathbf{A}} h_{k-1} + \overline{\mathbf{B}} p^{''}_{i,k},\\
    & p^{\diamond}_{i,k} = \mathbf{C} h_k + \mathbf{D} p^{''}_{i,k}.
\end{aligned}
\end{equation}
where $p^{''}_i=silu(p^{'}_i)$, $silu$ represents a gating activation function, $\overline{\mathbf{A}}$, $\overline{\mathbf{B}}$, $\mathbf{C}$, and $\mathbf{D}$ are discretized parameters of Selective State Space Model. The selective mechanism is achieved by designing $\overline{\mathbf{B}}$ and $\mathbf{C}$ matrices related to the input $p^{''}_{i,k}$. This is essentially a gating mechanism that filters redundant information, thereby attaining trusted fusion. The gating mechanism can effectively solve the problems of noise and redundant information in multi-view fusion as follows:
\begin{equation}
a_i=p^{\diamond}_i*silu(q_i), a_i \in \mathbb{R}^{Ml \times d^{'}},
\end{equation}
where $*$ represents the dot product, which multiplies the corresponding elements of the matrices of two branches. Then, we reduce the dimensionality of $a_i$ by $a^{'}_i=mlp^{3}(a_i)$.

\noindent\textbf{Convert Network.} Further, we employ a convert network to transform a fused sequence vector into a one-dimensional feature vector, as follows:
\begin{equation}
 u_i=rea^{4}(a^{'}_i), u_i \in \mathbb{R}^{Mld}.
\end{equation}

\subsection{Average-similarity Contrastive Learning}
This paper develops AsCL to solve the conflict problem in Contrastive Learning of deep multi-view clustering. We first calculate the similarity matrix of individual views for all samples as follows:
\begin{equation}
{S}^m_{i j}=\cos \left({z}^m_{i}, {z}^m_{j}\right),
\end{equation}
where $\cos$ is the function of cosine similarity. Then summing up the similarity matrices of all views and taking the average, we obtain
\begin{equation} \label{eq:sm}
S_{ij} = \frac{1}{M} \sum_{m=1}^{M} S_{ij}^m.
\end{equation}

AsCL unifies the dimensions of each view feature and the fused feature as follows:
\begin{equation}
{\hat{h}}_{i}=mlp^{4}(a^{''}_i),
\end{equation}
where the dimensionality of $a^{''}_i$ is reduced by the $mlp$ network.
We use the $mlp$ network to reduce dimensionality on each view feature $z^{m}_i$ in the same way,
\begin{equation}
{{h}^{m}_{i}}=mlp^{5,m}(z^{m}_i).
\end{equation}

The cosine distance is also utilized to calculate the similarity between fused presentation ${\hat{h}_i}$ and view-specific presentation ${h}^{m}_i$:
\begin{equation}
\label{eq:sim}
\begin{aligned}
C\left({\hat{h}}_{i}, {h}_i^{m}\right)= \cos ({\hat{h}}_{i}, {h}_i^{m}).
\end{aligned}
\end{equation}

The loss of Average-similarity Contrastive Learning is determined by the following:
\begin{equation}
\label{eq:lc}
\begin{aligned}
\mathcal{L}_{\mathrm{Asc}}=-\frac{1}{2 N} \sum_{i=1}^{N} \sum_{m=1}^{M} \log \frac{e^{\operatorname{C}\left({\hat{h}}_{i}, {h}_{i}^{m}\right) / \tau}}{\sum_{j=1}^{N} e^{(1-{S}_{ij})\operatorname{C}\left({\hat{h}}_{i}, {h}_{j}^{m}\right) / \tau}-e^{1 / \tau}},
\end{aligned}
\end{equation}
where $\tau$ represents the temperature coefficient. In Eq. (\ref{eq:sm}), ${S}_{ij}$ is calculated. The $\operatorname{C}\left({\hat{h}}_{i},{h}_{j}^{m}\right)$ in this equation increases with decreased ${S}_{ij}$ value. Stated differently, when the structural relationship \( S_{ij} \) between the \( i \)-th and \( j \)-th samples is low (i.e., they do not belong to the same cluster), their corresponding representations are inconsistent. Conversely, if the relationship is strong (indicating they are from the same cluster), their associated representations are consistent, leading to improved clustering results. The total loss is calculated as follows:
\begin{equation}
\label{zong}
{\cal L} = {{\cal L}_{\rm{Rec}}} + \lambda {{\cal L}_{\rm{Asc}}}.\\
\end{equation}

\begin{table}[!t]
\centering
\renewcommand\tabcolsep{6pt}
\setlength{\abovecaptionskip}{0.1cm}  
\caption{Description of the multi-view datasets.}
\begin{tabular}{ccccc} 
\toprule
Datasets    & Samples & Views & Clusters & View dimensions   \\ 
\midrule
Hdigit & 10000   & 2     & 10    & [784, 256]    \\
Cifar100 & 50000   & 3     & 100  &  [512,2048, 1024]   \\
Prokaryotic & 551     & 3     & 4   &  [438, 3, 393]    \\
Wiki & 2866     & 2     & 10   &  [128, 10]    \\
\bottomrule
\end{tabular}
\label{tab:Datasets}
\end{table}

\begin{table*}[ht]
\renewcommand\arraystretch{1}
\small
\setlength{\tabcolsep}{5.5pt}
\setlength{\abovecaptionskip}{0.1cm}  
\centering
\caption{Clustering result comparison for different datasets. The best results are bolded, and the second-best results are underlined.}
\begin{tabular}{c|ccc|ccc|ccc|ccc} 
\toprule
Datasets       & \multicolumn{3}{c|}{Hdigit}                           &\multicolumn{3}{c|}{Cifar100}          & \multicolumn{3}{c|}{Prokaryotic}                           & \multicolumn{3}{c}{Wiki}                           \\ 
\midrule
Metrics       & ACC           & NMI           & PUR           & ACC           & NMI           & PUR           & ACC           & NMI           & PUR           & ACC           & NMI           & PUR                          \\
\midrule
DEMVC \cite{xu2021deep}    & 0.3738 &0.3255 &0.4816 & 0.5048 &0.8343 &0.5177 & \underline{0.5245} &\underline{0.3079} &\underline{0.6969}          & 0.2544 &0.2409 &\underline{0.3126}                       \\
SiMVC \cite{trosten2021reconsidering}    & 0.7854 &0.6705 &0.7854    & 0.5795 &0.9225 &0.5869  & 0.5009 &0.1945 &0.6098                  & 0.2174 &0.0703 &0.2216                   \\
CoMVC \cite{trosten2021reconsidering}    & 0.9032 &0.8713 &0.9032  & \underline{0.6569} &\underline{0.9345} &\underline{0.6570} & 0.4138 &0.1883 &0.6697                  & 0.2694 &0.2624 &0.2903                       \\
MFLVC \cite{xu2022multi}    & 0.9257 &0.8396 &0.9257             & 0.1342 &0.0070 &0.1364    & 0.4301 &0.2216 &0.5989                  & \underline{0.3838} &\underline{0.2961} &0.2165                   \\
GCFAggMVC \cite{Yan_2023_CVPR}  & \underline{0.9730} &\underline{0.9274} &\underline{0.9730}    & 0.4370 &0.7718 &0.4783     & 0.4701 &0.1708 &0.5771                  & 0.1284 &0.0058 &0.1574                   \\ 
TMCN (Ours) & \textbf{0.9756} & \textbf{0.9341} & \textbf{0.9756} &\textbf{0.9853} &\textbf{0.9973} &\textbf{0.9897}  & \textbf{0.6715} &\textbf{0.4076} &\textbf{0.8094}        & \textbf{0.5691} & \textbf{0.5529} & \textbf{0.6354}  \\
\bottomrule
\end{tabular}
\label{tab:Clustering performance1}
\end{table*}

\subsection{Clustering module}
To achieve the clustering results for all samples, we use the k-means algorithm for the clustering module \cite{mackay2003information,bauckhage2015k,Yan_2023_CVPR}. In particular, the factorization of the learned fused representation $\widehat {\bf{H}}$ is as follows:
\begin{equation}
\label{clustering1}
\begin{array}{l}
\begin{array}{*{20}{c}}
{\mathop {\min }\limits_{{\bf{U,V}}} }&{{{\left\| {\widehat {\bf{H}} - {\bf{UV}}} \right\|}^2}}
\end{array}, \widehat {\bf{H}}=\{\hat{h}_1,...,\hat{h}_N\},\\
s.t.{\bf{U1}} = {\bf{1}},{\bf{U}} \ge {\bf{0}},
\end{array}
\end{equation}
where ${\bf{U}} \in \mathbb{R}^{N \times k} $ is matrix of cluster indicators; ${\bf{V}} \in \mathbb{R}^{k\times d^{''}}$ serves as the clustering center matrix.      


\section{Experiments}
\subsection{Experimental Settings}
We evaluate the proposed TMCN on four public multi-view datasets with different scales (see Table \ref{tab:Datasets}). For the evaluation metrics, three metrics, including accuracy (ACC), normalized mutual information (NMI), and Purity (PUR).

\textbf{Compared methods.} 
To evaluate the effectiveness of the proposed method, we compare the TMCN with Five state-of-the-art clustering methods, which are all deep methods (including DEMVC \cite{xu2021deep}, SiMVC \cite{trosten2021reconsidering}, CoMVC \cite{trosten2021reconsidering}, MFLVC \cite{xu2022multi} and GCFAggMVC \cite{Yan_2023_CVPR}).

\subsection{Experimental comparative results}
The comparative results with five methods by three evaluation metrics (ACC, NMI, PUR) on four benchmark datasets are presented as Table~\ref{tab:Clustering performance1}. 
The results show that the proposed TMCN is overall better than all the compared multi-view clustering methods by a large margin. Specifically, we obtain the following observations:
We compare five deep multi-view clustering methods (DEMVC, SiMVC, CoMVC, MFLVC, and GCFAggMVC) with the proposed method. On the Cifar100 dataset, our method outperforms the second-best method CoMVC by 32 percentage points in ACC. Similarly, on Prokaryotic, our proposed method performs better than the DEMVC method by 11 percentage points in terms of ACC metrics. Our proposed method also outperforms the baseline methods significantly in both NMI and PUR metrics.
The main reasons for these superior results come from two aspects: the TMFN module and the AsCL module.

\subsection{Ablation Study}
\par
We conducted an ablation study to evaluate each component of the proposed model.
\begin{table}[htb]
\small
\setlength{\tabcolsep}{8pt}
\setlength{\abovecaptionskip}{0.1cm}  
\centering
\caption{Ablation study on diverse datasets.}
\begin{tabular}{ccccc} 
\toprule
Datasets                     & Method      & ACC & NMI  & PUR \\ 
\midrule
\multirow{3}{*}{Hdigit} & No-TMFN & 0.9432 & 0.8985  & 0.9432 \\
                             & No-AsCL & 0.8764 & 0.7585 & 0.8764 \\
                             & TMCN       & \textbf{0.9756} & \textbf{0.9341} & \textbf{0.9756}  \\
\midrule
\multirow{3}{*}{Cifar100}         & No-TMFN & 0.9010  & 0.9811 & 0.9292 \\
                             & No-AsCL & 0.9582 & 0.9921 & 0.9695            \\
                             & TMCN & \textbf{0.9853}  & \textbf{0.9973}& \textbf{0.9897}   \\ 
\midrule
\multirow{3}{*}{Prokaryotic} & No-TMFN & 0.5771 & 0.3934  & 0.7915 \\
                             &  No-AsCL & 0.6134 & 0.3544 & 0.7623 \\
                             & TMCN        & \textbf{0.6715} & \textbf{0.4076} & \textbf{0.8094}  \\
\midrule
\multirow{3}{*}{Wikipedia} & No-TMFN & 0.5049 & 0.5021  & 0.5666 \\
                             &  No-AsCL & 0.5478 & 0.5354 & 0.6151 \\
                             & TMCN  & \textbf{0.5691} & \textbf{0.5529} & \textbf{0.6354}  \\
\bottomrule
\end{tabular}
\label{tab:Ablation}
\end{table}
\begin{figure}[htbp]
\centering
\subfigure[] { 
\includegraphics[width=1.75in]{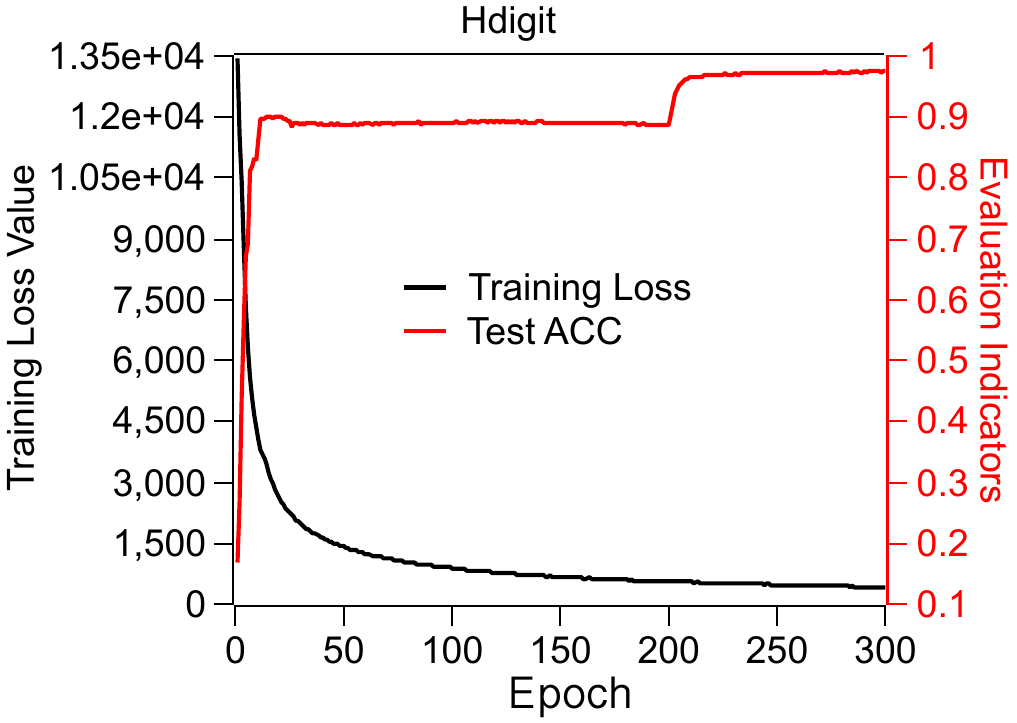}
}\hspace{0.1cm}
\subfigure[] { 
\includegraphics[width=1.35in]{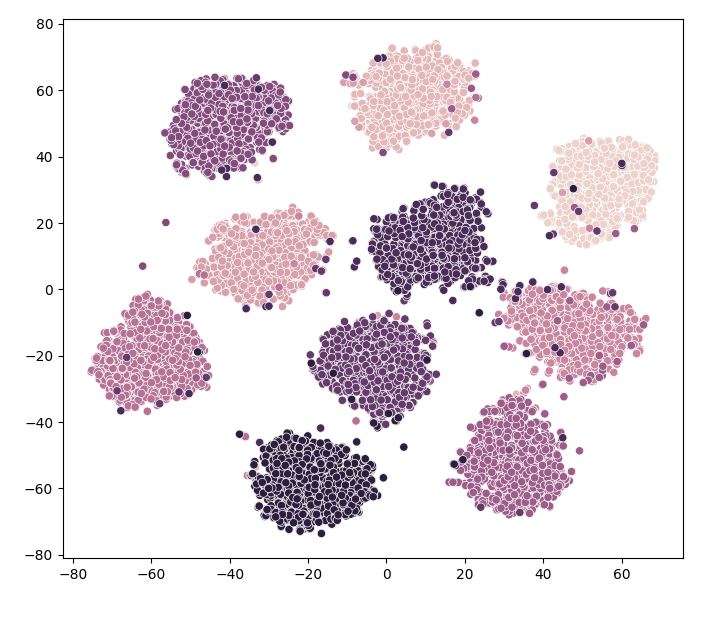}
}
\caption{The convergence analysis and visualization analysis on Hdigit.}
\label{fig:Visualization and convergence} 
\end{figure}

\begin{figure}[htbp]
\centering
\subfigure[] { 
\includegraphics[width=0.41\columnwidth]{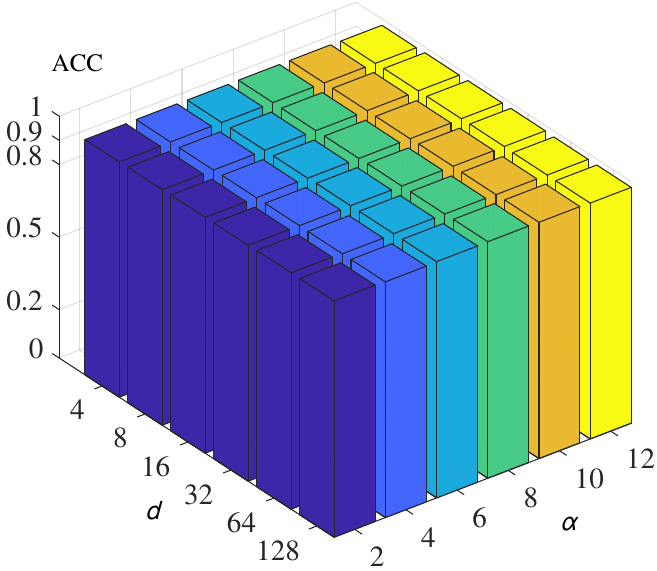}
}\hspace{0.25cm}
\subfigure[] { 
\includegraphics[width=0.412\columnwidth]{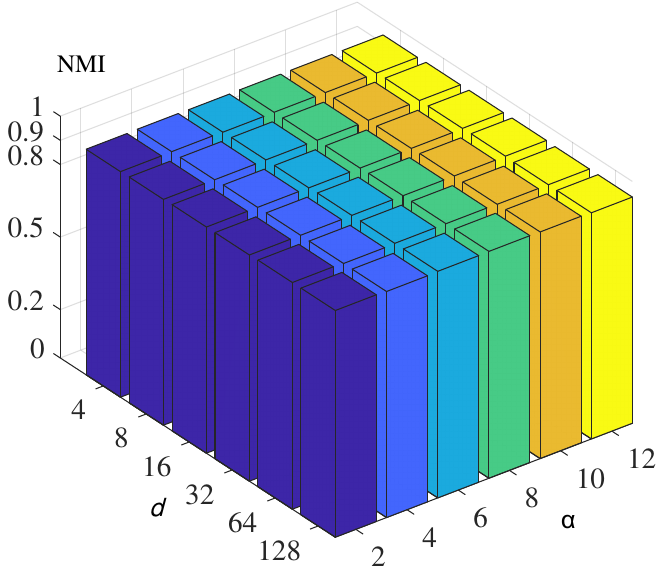}
}
\caption{The parameter analysis on Hdigit.}
\label{fig:hyper} 
\end{figure}

\textbf{Effectiveness of TMFN module.} The fused representation is set to $\mathbf{Z}$, which is the concatenation of all view-specific representations. The network is represented as ``No-TMFN". Table \ref{tab:Ablation} illustrates that, in the ACC term, the results of No-TMFN are $3.24$, $8.43$, $9.44$, and $6.42$ percent less than those of our method. The concatenated representation $\mathbf{Z}$ is not conducive to clustering since it contains much noise information. The TMFN thoroughly explores the selective mechanism of the Mamba network, effectively mitigating the disruptive effects of noise and redundancy across diverse views. The results demonstrate that the TMFN module significantly improves multi-view clustering performance.

\textbf{Validity of AsCL module.} According to Table \ref{tab:Ablation}, the results of No-AsCL are lower than those of the TMCN method by 9.92, 2.71, 5.81, and 2.13 percent in ACC term. Our fused representation of multiple views is improved by the similarity of view presentation from the same cluster, rather than simply the same sample. AsCL can effectively alleviate the conflict of samples of the same cluster in contrastive learning. Therefore, it enhances the performance of deep multi-view clustering.

\subsection{Convergence, Visualization, and Parameter Analysis.} 
To verify the convergence, we plot the objective values and evaluation metric values through iterations in Figure \ref{fig:Visualization and convergence}. It can be observed that the objective value monotonically decreases until convergence. The value of ACC first increases gradually with iteration and then fluctuates in a narrow range. These results all confirm the convergence of TMCN. In addition, to further verify the effectiveness of the proposed TMCN,  we visualize fused representations after convergence by the t-SNE method \cite{van2008visualizing} in Figure \ref{fig:Visualization and convergence}. Figure \ref{fig:hyper} shows the clustering results of the proposed TMCN  are insensitive to both $d$ and $\alpha$ in the range 4 to 128, and the range 2 to 12, respectively. 

\section{Conclusion and Future work}
This study introduces the TMCN framework, designed to facilitate trusted fusion for multi-view clustering. The TMFN module, which leverages the selective mechanism of the Mamba network, is proposed for the multi-view trusted fusion. Additionally, AsCL module is crafted to rectify the inconsistency in the representation space among samples within clusters, further enhancing trusted fusion. Experimental results conclusively demonstrate the exceptional performance of TMCN over state-of-the-art methods in clustering tasks.
\vspace{-9pt}

\section*{Acknowledgment}
This research is supported by the National Key R\&D Program of China (Grant No.2024YFB3213400), the Anhui Provincial Natural Science Foundation (Grant No.2408085QF214), the Fundamental Research Funds for the Central Universities (Grant No. WK2100000045), and the Opening Project of the State Key Laboratory of General Artificial Intelligence(Grant No. SKLAGI2024OP10, Grant No.SKLAGI2024OP11), the Young Scientists Fund of the National Natural Science Foundation of China (Grant No. 62306286), and the National Key R\&D Program of China (2022YFB4500405).


\bibliographystyle{IEEEtran}  
\bibliography{ref} 

\begin{thebibliography}{10}
\providecommand{\url}[1]{#1}
\csname url@samestyle\endcsname
\providecommand{\newblock}{\relax}
\providecommand{\bibinfo}[2]{#2}
\providecommand{\BIBentrySTDinterwordspacing}{\spaceskip=0pt\relax}
\providecommand{\BIBentryALTinterwordstretchfactor}{4}
\providecommand{\BIBentryALTinterwordspacing}{\spaceskip=\fontdimen2\font plus
\BIBentryALTinterwordstretchfactor\fontdimen3\font minus
  \fontdimen4\font\relax}
\providecommand{\BIBforeignlanguage}[2]{{%
\expandafter\ifx\csname l@#1\endcsname\relax
\typeout{** WARNING: IEEEtran.bst: No hyphenation pattern has been}%
\typeout{** loaded for the language `#1'. Using the pattern for}%
\typeout{** the default language instead.}%
\else
\language=\csname l@#1\endcsname
\fi
#2}}
\providecommand{\BIBdecl}{\relax}
\BIBdecl

\bibitem{chen2024end}
L.~Chen, P.~Wu, K.~Chitta, B.~Jaeger, A.~Geiger, and H.~Li, ``End-to-end
  autonomous driving: Challenges and frontiers,'' \emph{IEEE Transactions on
  Pattern Analysis and Machine Intelligence}, 2024.

\bibitem{zou2023dpnet}
X.~Zou, C.~Tang, X.~Zheng, Z.~Li, X.~He, S.~An, and X.~Liu, ``Dpnet: Dynamic
  poly-attention network for trustworthy multi-modal classification,'' in
  \emph{Proceedings of the 31st ACM International Conference on Multimedia},
  2023, pp. 3550--3559.

\bibitem{chao2021survey}
G.~Chao, S.~Sun, and J.~Bi, ``A survey on multiview clustering,'' \emph{IEEE
  transactions on artificial intelligence}, vol.~2, no.~2, pp. 146--168, 2021.

\bibitem{zou2023inclusivity}
X.~Zou, C.~Tang, X.~Zheng, K.~Sun, W.~Zhang, and D.~Ding, ``Inclusivity induced
  adaptive graph learning for multi-view clustering,'' \emph{Knowledge-Based
  Systems}, vol. 267, p. 110424, 2023.

\bibitem{xiao2024dual}
Y.~Xiao, D.~Yang, J.~Li, X.~Zou, H.~Zhou, and C.~Tang, ``Dual alignment feature
  embedding network for multi-omics data clustering,'' \emph{Knowledge-Based
  Systems}, p. 112774, 2024.

\bibitem{yang2024trustworthy}
W.~Yang, M.~Wang, C.~Tang, X.~Zheng, X.~Liu, and K.~He, ``Trustworthy
  multi-view clustering via alternating generative adversarial representation
  learning and fusion,'' \emph{Information Fusion}, vol. 107, p. 102323, 2024.

\bibitem{dang2024exploring}
Y.~Dang, M.~Gao, Y.~Yan, X.~Zou, Y.~Gu, A.~Liu, and X.~Hu, ``Exploring response
  uncertainty in mllms: An empirical evaluation under misleading scenarios,''
  \emph{arXiv preprint arXiv:2411.02708}, 2024.

\bibitem{he2023multispectral}
X.~He, C.~Tang, X.~Zou, and W.~Zhang, ``Multispectral object detection via
  cross-modal conflict-aware learning,'' in \emph{Proceedings of the 31st ACM
  International Conference on Multimedia}, 2023, pp. 1465--1474.

\bibitem{xu2025hstrans}
K.~Xu, M.~Wang, X.~Zou, J.~Liu, A.~Wei, J.~Chen, and C.~Tang, ``Hstrans:
  Homogeneous substructures transformer for predicting frequencies of drug-side
  effects,'' \emph{Neural Networks}, vol. 181, p. 106779, 2025.

\bibitem{zou2024dai}
X.~Zou, X.~He, X.~Zheng, W.~Zhang, J.~Chen, and C.~Tang, ``Dai-net: Dual
  adaptive interaction network for coordinated medication recommendation,''
  \emph{IEEE Journal of Biomedical and Health Informatics}, 2024.

\bibitem{zou2023hierarchical}
X.~Zou, C.~Tang, W.~Zhang, K.~Sun, and L.~Jiang, ``Hierarchical attention
  learning for multimodal classification,'' in \emph{2023 IEEE International
  Conference on Multimedia and Expo (ICME)}.\hskip 1em plus 0.5em minus
  0.4em\relax IEEE, 2023, pp. 936--941.

\bibitem{zhu:23}
J.~Zhu, X.~Ruan, Y.~Cheng, Z.~Huang, Y.~Cui, and L.~Zeng, ``Deep metric
  multi-view hashing for multimedia retrieval,'' in \emph{2023 IEEE
  International Conference on Multimedia and Expo (ICME)}, 2023, pp.
  1955--1960.

\bibitem{zhu:24}
J.~Zhu, P.~Hu, B.~Li, and Y.~Zhou, ``Fast metric multi-view hashing for
  multimedia retrieval,'' \emph{Information Fusion}, vol. 103, p. 102130, 2024.

\bibitem{zhu:25}
J.~Zhu, Y.~Cui, Z.~Huang, X.~Li, L.~Liu, L.~Zeng, and L.-R. Dai, ``Adaptive
  confidence multi-view hashing for multimedia retrieval,'' in \emph{ICASSP
  2024 - 2024 IEEE International Conference on Acoustics, Speech and Signal
  Processing (ICASSP)}, 2024, pp. 7900--7904.

\bibitem{zhu:26}
J.~Zhu, W.~Cheng, Y.~Cui, C.~Tang, Y.~Dai, Y.~Li, and L.~Zeng, ``Central
  similarity multi-view hashing for multimedia retrieval,'' in \emph{Web and
  Big Data}.\hskip 1em plus 0.5em minus 0.4em\relax Springer Nature Singapore,
  2024, pp. 486--500.

\bibitem{zou2024look}
X.~Zou, Y.~Wang, Y.~Yan, S.~Huang, K.~Zheng, J.~Chen, C.~Tang, and X.~Hu,
  ``Look twice before you answer: Memory-space visual retracing for
  hallucination mitigation in multimodal large language models,'' \emph{arXiv
  preprint arXiv:2410.03577}, 2024.

\bibitem{zhu:27}
J.~Zhu, Z.~Huang, L.~Liu, C.~Tang, and L.-R. Dai, ``Boosted curriculum
  multi-view hashing for multimedia retrieval,'' \emph{IEEE Signal Processing
  Letters}, vol.~31, pp. 2065--2069, 2024.

\bibitem{du2021deep}
G.~Du, L.~Zhou, Y.~Yang, K.~L{\"u}, and L.~Wang, ``Deep multiple
  auto-encoder-based multi-view clustering,'' \emph{Data Science and
  Engineering}, vol.~6, no.~3, pp. 323--338, 2021.

\bibitem{abavisani2018deep}
M.~Abavisani and V.~M. Patel, ``Deep multimodal subspace clustering networks,''
  \emph{IEEE Journal of Selected Topics in Signal Processing}, vol.~12, no.~6,
  pp. 1601--1614, 2018.

\bibitem{zhou2020end}
R.~Zhou and Y.-D. Shen, ``End-to-end adversarial-attention network for
  multi-modal clustering,'' in \emph{Proceedings of the IEEE/CVF conference on
  computer vision and pattern recognition}, 2020, pp. 14\,619--14\,628.

\bibitem{xu2021deep}
J.~Xu, Y.~Ren, G.~Li, L.~Pan, C.~Zhu, and Z.~Xu, ``Deep embedded multi-view
  clustering with collaborative training,'' \emph{Information Sciences}, vol.
  573, pp. 279--290, 2021.

\bibitem{trosten2021reconsidering}
D.~J. Trosten, S.~Lokse, R.~Jenssen, and M.~Kampffmeyer, ``Reconsidering
  representation alignment for multi-view clustering,'' in \emph{Proceedings of
  the IEEE/CVF Conference on Computer Vision and Pattern Recognition}, 2021,
  pp. 1255--1265.

\bibitem{xu2022multi}
J.~Xu, H.~Tang, Y.~Ren, L.~Peng, X.~Zhu, and L.~He, ``Multi-level feature
  learning for contrastive multi-view clustering,'' in \emph{Proceedings of the
  IEEE/CVF Conference on Computer Vision and Pattern Recognition}, 2022, pp.
  16\,051--16\,060.

\bibitem{hershey2007approximating}
J.~R. Hershey and P.~A. Olsen, ``Approximating the kullback leibler divergence
  between gaussian mixture models,'' in \emph{2007 IEEE International
  Conference on Acoustics, Speech and Signal Processing-ICASSP'07},
  vol.~4.\hskip 1em plus 0.5em minus 0.4em\relax IEEE, 2007, pp. IV--317.

\bibitem{tang2022deep}
H.~Tang and Y.~Liu, ``Deep safe incomplete multi-view clustering: Theorem and
  algorithm,'' in \emph{Proceedings of the 39th International Conference on
  Machine Learning}, 2022, pp. 162:21\,090--21\,110.

\bibitem{xu2022deep}
J.~Xu, C.~Li, Y.~Ren, L.~Peng, Y.~Mo, X.~Shi, and X.~Zhu, ``Deep incomplete
  multi-view clustering via mining cluster complementarity,'' in
  \emph{Thirty-Six AAAI conference on artificial intelligence}, 2022, pp.
  8761--8769.

\bibitem{Albert:28}
A.~Gu and T.~Dao, ``Mamba: Linear-time sequence modeling with selective state
  spaces,'' \emph{CoRR}, vol. abs/2312.00752, 2023.

\bibitem{song2018self}
J.~Song, H.~Zhang, X.~Li, L.~Gao, M.~Wang, and R.~Hong, ``Self-supervised video
  hashing with hierarchical binary auto-encoder,'' \emph{IEEE Transactions on
  Image Processing}, vol.~27, no.~7, pp. 3210--3221, 2018.

\bibitem{hinton2006reducing}
G.~E. Hinton and R.~R. Salakhutdinov, ``Reducing the dimensionality of data
  with neural networks,'' \emph{science}, vol. 313, no. 5786, pp. 504--507,
  2006.

\bibitem{Yan_2023_CVPR}
W.~Yan, Y.~Zhang, C.~Lv, C.~Tang, G.~Yue, L.~Liao, and W.~Lin, ``Gcfagg: Global
  and cross-view feature aggregation for multi-view clustering,'' in
  \emph{Proceedings of the IEEE/CVF Conference on Computer Vision and Pattern
  Recognition (CVPR)}, June 2023, pp. 19\,863--19\,872.

\bibitem{mackay2003information}
D.~J. MacKay, D.~J. Mac~Kay \emph{et~al.}, \emph{Information theory, inference
  and learning algorithms}.\hskip 1em plus 0.5em minus 0.4em\relax Cambridge
  university press, 2003.

\bibitem{bauckhage2015k}
C.~Bauckhage, ``K-means clustering is matrix factorization,'' \emph{arXiv
  preprint arXiv:1512.07548}, 2015.

\bibitem{van2008visualizing}
L.~Van~der Maaten and G.~Hinton, ``Visualizing data using t-sne.''
  \emph{Journal of machine learning research}, vol.~9, no.~11, 2008.

\end{thebibliography}

\end{document}